\documentclass{article} %
\usepackage{iclr2020_conference,times}
\pdfoutput=1

\usepackage{amsmath,amsfonts,bm}

\def\eqref#1{equation~\ref{#1}}

\def\1{\bm{1}}

\DeclareMathAlphabet{\mathsfit}{\encodingdefault}{\sfdefault}{m}{sl}
\SetMathAlphabet{\mathsfit}{bold}{\encodingdefault}{\sfdefault}{bx}{n}

\usepackage{hyperref}
\usepackage{url}
\usepackage{multirow}

\title{Evaluating Amharic Machine Translation}

\author{Asmelash Teka Hadgu, Adam Beaudoin, Abel Aregawi \\
\texttt{\{asme,adam,abel\}@lesan.ai}
}

\iclrfinalcopy %
\begin{document}

\maketitle

\begin{abstract}
Machine translation (MT) systems are now able to provide very accurate results for high resource language pairs. However, for many low resource languages, MT is still under active research. In this paper, we develop and share a dataset~\footnote{\url{https://zenodo.org/record/3734260}} to automatically evaluate the quality of MT systems for Amharic. We compare two commercially available MT systems that support translation of Amharic to and from English to assess the current state of MT for Amharic. The BLEU score results show that the results for Amharic translation are promising but still low. We hope that this dataset will be useful to the research community both in academia and industry as a benchmark to evaluate Amharic MT systems.
\end{abstract}

\section{Introduction}

Machine translation (MT) has a lot of applications in different domains such as consumer reviews for marketplaces~\citep{guha2014machine}, insights and sentiment analysis for social media posts~\citep{balahur2012multilingual}, improving human translation speed~\citep{koehn2009interactive} and high volume content translation for web browsers. Implementation of MT has been shown to increase international trade~\citep{brynjolfsson2019does}. While there are many applications for MT, the quality of commercial systems available on the market varies greatly from system to system and language to language.

One of the challenges of building effective MT systems is quantifying quality improvements. E.g., an MT system trained on religious texts for training and testing may seem to have good results but in reality may not reflect quality improvement. This could be a problem with (i) data leakage (having similar/same sentences between train and test) or (ii) domain mismatch (different domains between train and test). A robust and reliable, diverse benchmark is vital to determine the quality of MT systems. In this paper, we contribute a dataset~\citep{am_mt_3734260} toward that effort for Amharic – the official language of Ethiopia.

\section{Dataset}

\cite{abate2018parallel} shared parallel corpus to build MT systems for seven Ethiopian language pairs. The design goal of an evaluation set is different in that it is designed to test different use-cases of an MT system. Hence, the evaluation dataset should cover broad categories where MT systems are applicable. We identified news, general knowledge text, short messages and everyday conversational texts to cover these aspects. 

To evaluate an MT system for Amharic to and from English, we need to collect two types of sources corresponding to the two languages. Broadly, we want the source Amharic sentences to cover local events and the source English sentences to describe global events.

\paragraph{Collection and Preprocessing of Amharic Sources}
Our requirement for Amharic sources was to cover local content.
\begin{itemize}
    \item News: we gathered news headlines from Amharic news websites including: \href{https://www.press.et/}{Ethiopian Press Agency}, \href{https://fanabc.com/}{Fana Broadcasting Corporate}, \href{https://www.ethiopianreporter.com/}{The Ethiopian Reporter}, \href{http://www.waltainfo.com/}{Walta Information Center} and \href{https://www.bbc.com/amharic}{Amharic BBC} across different categories such as: politics, business and economy, health, sport etc.
    \item Wikipedia: We took the Wikipedia dump for Amharic~\footnote{\url{https://dumps.wikimedia.org/other/cirrussearch/20200120/amwiki-20200120-cirrussearch-content.json.gz}}, extracted sentences, ran language identification for Amharic and dropped sentences that do not have at least two tokens. We then randomly selected sentences for translation by humans.
    \item Twitter: We collected Amharic tweets by searching for Amharic stop-words using~\cite{tas_2605413} a wrapper around the Twitter advanced search~\footnote{\url{https://twitter.com/search-advanced}}. Similar to Wikipedia, we performed language identification, and dropped tweets that do not have at least two tokens.
    \item Conversational: For conversational type of text, we used Amharic native speakers to get Amharic source sentences and their corresponding translations.
\end{itemize}

\paragraph{Collection and Preprocessing of English Sources}
For English sources, our requirement was that the content should be of global interest. We found Wikipedia current events portal~\footnote{\url{https://en.wikipedia.org/wiki/Portal:Current_events}}, meets this requirement.
The Wikipedia current events portal contains news listed on a daily basis with a link to the corresponding background articles. Of primary importance are events. However, it also contains trends and developments.
These news items contain different categories, such as: armed conflicts, arts and culture, business and economy, disasters and accidents, health and environment, international relations, law and crime, politics and
elections, science and technology, sports. In our collection the time span ranges from 2013 to 2020 inclusive.
\begin{itemize}
    \item  News: For news headlines, we took a sample of news headlines across the different categories for each year from the Wikipedia current events portal.
    \item  Twitter: Wikipedia events portal already contains named entities in each news headline. We took these entities and the corresponding date as query parameters and performed a search on Twitter advanced search using~\cite{tas_2605413} which is available on Github~\footnote{\url{https://github.com/asmelashteka/twitteradvancedsearch}}.
    \item Wikipedia: Similar to the Amharic Wikipedia, we used the English Wikipedia dump~\footnote{\url{https://dumps.wikimedia.org/other/cirrussearch/20200120/enwiki-20200120-cirrussearch-content.json.gz}}, extracted sentences, ran language identification for English and dropped sentences that do not have at least two tokens.
    \item Conversational: We collected common phrases and expressions from native speakers and websites for spoken English that were used as proxy for every day conversational English.
\end{itemize}
The overall translation process of these source sentences is described as follows.

\paragraph{Annotation}
To generate a diverse dataset of conversational, tweets, Wikipedia and news content a job  advertisement for a short text translator was posted to a Telegram group. 198 applicants inquired for the position and after screening the quality of their response, 124 applicants were sent a 10 sentence Amharic sample of either Wikipedia content or news headlines. The work was reviewed and given a score to evaluate and select translations only if they score above a threshold. Scores of bad, ok, good, very good and perfect were assigned. Any content below a good level of quality was not used. After reviewing the initial sample, 28 applicants were given 50 lines of either Wikipedia content or news headlines to translate. Their work was screened and post edited to ensure a high level of translation quality. Successful applicants were asked to provide conversational text and were given English news, tweets and Wikipedia text to translate as well. 
Tweet translation and post editing was completed using a similar process. Translating tweets is challenging in that they are very informal, have code-mixing and social media specific features such as @mentions, hashtags and embedded URLs. Table~\ref{data-ground-truth} shows the domain and number of bitext in the evaluation dataset.

\begin{table}[ht]
\caption{Domain and number of bitext used for evaluation}
\label{data-ground-truth}
\begin{center}
\begin{tabular}{r|c|c|c|c|c}
Source & News & Wikipedia & Twitter & Conversational & total bitext \\
\hline
Amharic & 383 & 257 & 203 & 154 & 997 \\
English & 371 & 499 & 641 & 404 & 1915 \\
\end{tabular}
\end{center}
\end{table}

\section{Evaluation and Results}

Two commercial systems were used to evaluate the current state of MT for Amharic. These are Google Translate~\footnote{\url{https://translate.google.com/}} and Yandex
Translate~\footnote{\url{https://translate.yandex.com/}}. Google uses neural MT~\citep{wu2016google} which has made significant advances in the quality of MT systems for many languages. Yandex Translate uses a hybrid system of neural MT and statistical MT~\footnote{\url{https://tech.yandex.com/translate/doc/dg/concepts/how-works-machine-translation-docpage/}}. Both services provide APIs to access their system. A user can send a string and get back the corresponding translation. The systems were accessed on 14\textsuperscript{th} February 2020 for Amharic source (to evaluate Am $\rightarrow$ En) and 30\textsuperscript{th} March 2020 for English source (to evaluate En $\rightarrow$ Am) respectively. We used the SacreBLEU~\citep{post-2018-call} implementation to evaluate the results. We tried out different tokenizers. We report the results when tokenization is set to `none'~\footnote{BLEU+case.mixed+numrefs.1+smooth.exp+tok.none+version.1.4.6}. The default tokenization on SacreBLEU, 13a~\footnote{BLEU+case.mixed+numrefs.1+smooth.exp+tok.13a+version.1.4.6}, inflates the result of the tweet translation. This is because it tokenizes URLs to many sub-tokens that overwhelm the sentence tokens. We chose tokenization `none' for consistency but we also report the difference when using the `13a' tokenizer for News, Wikipedia and Conversational domains. 

Table~\ref{res-BLEU} shows the BLEU scores for Google Translate and Yandex Translate. In all domains Google Translate is by far better than Yandex translate. Google Translate does better on news headlines and conversational type texts than on Wikipedia and Twitter. Both systems perform worse when translating from English to Amharic. This is a key finding for researchers in low resource machine translation. There is a huge opportunity to improve these systems and allow for more communities to benefit from this technology.

\begin{table}[ht]
\caption{BLEU score for Amharic to English and English to Amharic translation using two commercial MT systems.}
\label{res-BLEU}
\begin{center}
\begin{tabular}{r|r|c|c|c|c|c}
Direction & Service & News & Wikipedia & Twitter & Conversational & All combined\\
\hline
\multirow{2}{*}{Am $\rightarrow$ En} & Google Translate & 30.8 \tiny{+ 0.8} & 18.3 \tiny{+ 3.6} & 19.2 & 30.5 \tiny{+ 1.6} & 23.2 \\
                                     & Yandex Translate & 1.2  \tiny{+ 1.3} & 1.5 \tiny{+ 2.6}  & 8.6  & 2.4 \tiny{+ 2.5} & 4.8 \\ 
\hline
\multirow{2}{*}{En $\rightarrow$ Am} & Google Translate & 13.7 \tiny{+ 0.4} & 8.5 \tiny{+ 0.7} & 7.6 & 4.8 \tiny{+ 2.8} & 9.6\\
                                     & Yandex Translate & 0.4 \tiny{+ 0.1}  & 0.3 \tiny{+ 0.6} & 2.3 & 0.4 \tiny{+ 0.3} & 1.3 \\

\end{tabular}
\end{center}
\end{table}

\section{Conclusion}

In this work, we provided an evaluation dataset to asses the quality of machine translation systems for Amharic. The dataset is diverse, containing text from news headlines, Wikipedia, Twitter and conversational types. 
We also evaluated current state-of-the-art MT systems for the English Amharic language pair. We found that while Google Translate does well on Amharic to English translation, the English to Amharic translation from both systems is poor. In future work, we will continue to increase the amount and variety to the dataset. We believe that such a benchmark dataset is valuable for the research community to compare and evaluate their systems.

\bibliography{iclr2020_conference}
\bibliographystyle{iclr2020_conference}

\end{document}